\documentclass{article}

     \usepackage[nonatbib, final]{neurips_2020}

\usepackage[utf8]{inputenc} % allow utf-8 input
\usepackage[T1]{fontenc}    % use 8-bit T1 fonts
\usepackage{hyperref}       % hyperlinks
\usepackage{url}            % simple URL typesetting
\usepackage{booktabs}       % professional-quality tables
\usepackage{amsfonts}       % blackboard math symbols
\usepackage{nicefrac}       % compact symbols for 1/2, etc.
\usepackage{microtype}      % microtypography
\usepackage{algorithm}
\usepackage{algpseudocode}
\usepackage{graphicx}
\usepackage[style=apa,backend=biber,citestyle=numeric]{biblatex}
\DeclareFieldFormat{labelnumberwidth}{\mkbibbrackets{#1}}

\defbibenvironment{bibliography}
  {\list
     {\printtext[labelnumberwidth]{%
      \printfield{labelprefix}%
      \printfield{labelnumber}}}
     {\setlength{\labelwidth}{\labelnumberwidth}%
      \setlength{\leftmargin}{\labelwidth}%
      \setlength{\labelsep}{\biblabelsep}%
      \addtolength{\leftmargin}{\labelsep}%
      \setlength{\itemsep}{\bibitemsep}%
      \setlength{\parsep}{\bibparsep}}%
      }
  {\endlist}
  {\item}

\title{Seeded iterative clustering for histology region identification}

\author{%
Eduard Chelebian  \\
Department of Information Technology\\
Uppsala University, SciLifeLab\\
Uppsala, Sweden \\
\texttt{eduard.chelebian@it.uu.se} \\
\And
Francesco Ciompi \\
Department of Pathology \\ 
Radboud University Medical Center\\
Nijmegen, The Netherlands \\
\texttt{francesco.ciompi@radboudumc.nl} \\
\AND
Carolina Wählby  \\
Department of Information Technology\\
Uppsala University, SciLifeLab\\
Uppsala, Sweden \\
\texttt{carolina.wahlby@it.uu.se} \\
}

\begin{document}

\maketitle

\begin{abstract}

Annotations are necessary to develop computer vision algorithms for histopathology, but dense annotations at a high resolution are often time-consuming to make. Deep learning models for segmentation are a way to alleviate the process, but require large amounts of training data, training times and computing power. To address these issues, we present \emph{seeded iterative clustering} to produce a coarse segmentation densely and at the whole slide level. The algorithm uses precomputed representations as the clustering space and a limited amount of sparse interactive annotations as seeds to iteratively classify image patches. We obtain a fast and effective way of generating dense annotations for whole slide images and a framework that allows the comparison of neural network latent representations in the context of transfer learning. 

\end{abstract}

\section{Introduction}

Deep learning segmentation models are already being used for easing the burden of cancer histopathology whole slide image (WSI) annotation \cite{van2021deep}. However, supervised training requires annotated images, often long training times and computational resources such as graphical processing units (GPU). To overcome these bottlenecks, approaches such as active learning \cite{ren2021survey} are gaining popularity, as they reduce the time needed to perform the segmentation while retaining high performance via expert supervision. One strategy is to have pathologists annotate a small part of the image, either by clicking or scribbling around the regions of interest. Then, features are extracted from those sparsely annotated regions and the dense segmentation masks are obtained by retraining segmentation models. This strategy, however, requires both interactive and GPU support at run time.

For this proof-of-concept, we instead propose seeded interactive clustering (SIC) to decouple the problem by (1) precomputing the latent network representations, which does not require a GPU at run time so it can be done remotely in a computing cluster without interaction and, once the embeddings are obtained, (2)  perform interactive clustering locally that does not involve any further training and is relatively cheaper computationally.

\subsection{Previous work}

Different authors have approached this problem by creating interactive tools based on extracting classical features \cite{hollandi2020annotatorj, arzt2022labkit} or on deep learning approaches \cite{lindvall2020tissuewand, jaber2021deep, miao2021quick}. Some of these approaches are intended for cell detection on smaller images, and therefore are harder to scale to bigger WSI. The approaches that are meant for digital pathology usually require a GPU while performing the interactive annotations, which may not always be available.

The re-purposing of convolutional neural networks (CNN) as feature extractors was proposed very early after the first models trained on large datasets were made avaialable \cite{zeiler2014visualizing, donahue2014decaf, oquab2014learning, sharif2014cnn}. Since then, using model architectures such as ResNet-50 \cite{he2016deep} trained on ImageNet has become common practice for extracting features and performing transfer learning, even if some authors argue that these models may not capture all the relevant histology features \cite{chen2022self}.

Recent work also proposes using self-supervised learning approaches for obtaining the embeddings \cite{chen2020simple, ciga2022self}, under the assumption that the feature space would be more representative of the tissue differences. However comparing different embeddings is not a trivial task and is influenced by many factors \cite{kornblith2019similarity, neyshabur2020being}. 

Finally, image-set clustering is a field where all these concepts, including the choice of architecture and layer for feature extraction, or the clustering algorithms, has been extensively studied \cite{guerin2021combining}. 

\begin{figure}[h!]
  \centering
    \includegraphics[width=1\linewidth]{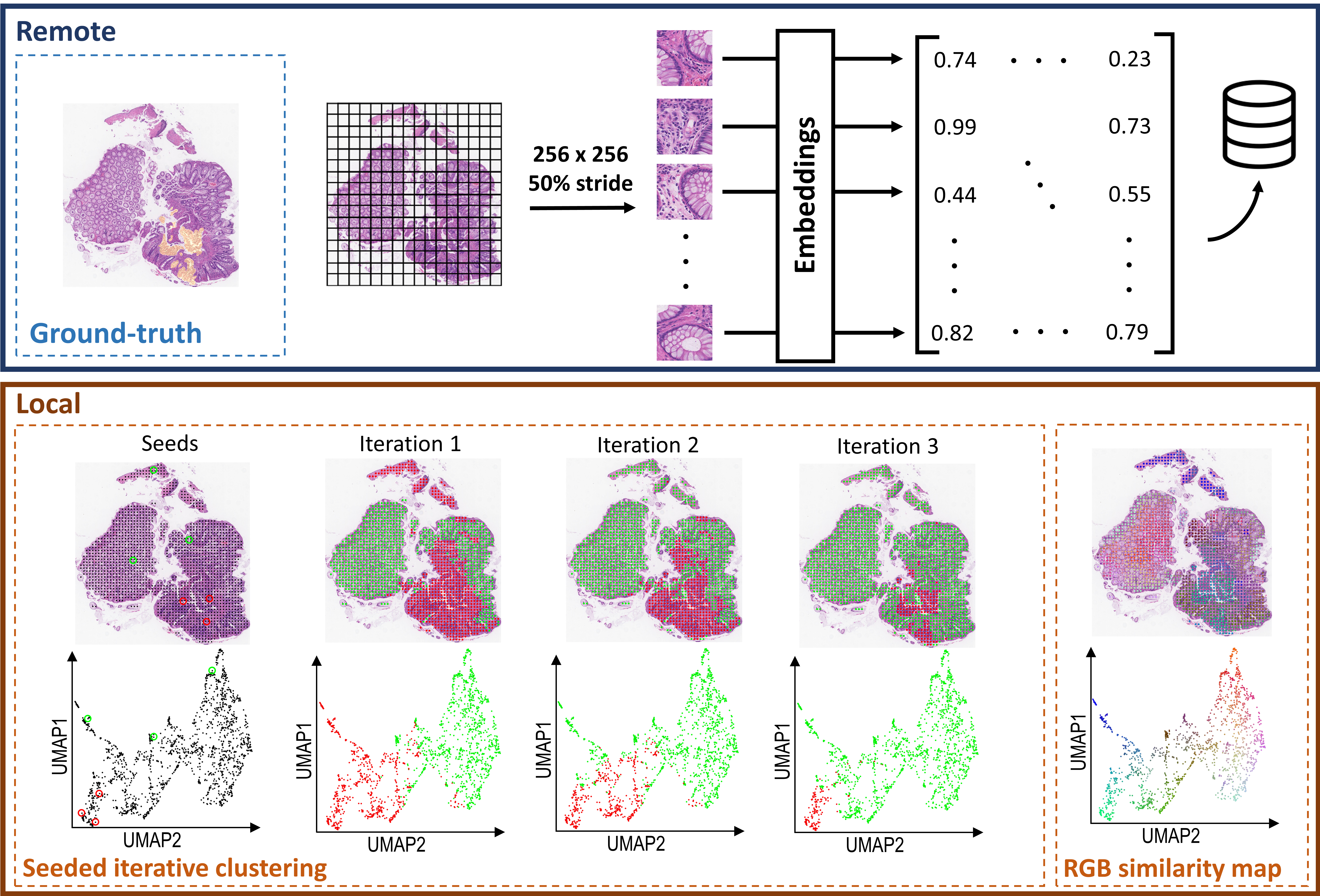}
  \caption{Visualization of the seeded interactive clustering process. The \textit{Remote} box includes the ground truth (left) and illustrates how tissue patch embeddings are extracted (right). The \textit{Local} box, from left to right: the top row shows a tissue sample, where seeds (red and green markers) are manually placed on the tissue image. The bottom row shows the embedding space location of the corresponding tissue patches, where black dots are patches not yet assigned to a category. As the SIC iterations continue, the patches are re-clustered according to embedding similarity. Far right: dots representing patches in UMAP space are colored according to their position in RGB color space, and re-mapped to their coordinates in tissue space. Note that the clustering is made on the full embedding space and the reduced space is only intended for visualization purposes. Visualizations created using TissUUmaps \cite{pielawski2022tissuumaps}.}
  \label{fig1}
\end{figure}

\section{Method}

\subsection{Seeded iterative clustering}

The proposed pipeline is presented in Figure \ref{fig1} and has two parts: one remote and one local. The first remote part of the process starts by dividing the image into smaller patches. The size of the patches, the stride between them and the resolution of choice defines the granularity of the final tissue clustering. In this case, we decided to extract $256\times256$ patches with $50\%$ overlap at $20\times$ resolution. Next, each of the patches are embedded using the pretrained model of choice by extracting the last layer latent representation. Finally, we store the matrix of $number\:of\:patches \times embedding\:size$ together with the center coordinates of each patch to represent the complete WSI.  

The second part of the pipeline can be run locally. The seeded iterative clustering (SIC) approach, presented in Algorithm \ref{alg:cap}, divides the previously generated patch embedding space in a binary fashion, guided by small seeds -or sparse annotations- selected by the user in representative regions. Then seeded clustering \cite{basu2002semi} is performed iteratively until a score relative to the sparse annotations does not improve. We used F1-score, but any classification metric could be used to optimize the algorithm. The seeded KMeans algorithm only differs from the original KMeans in the first assignment of cluster centers, which are not initialized randomly but based on the seeds. An additional step of conventional KMeans is added as it had shown to improve the performance when the tumor region of the WSI was not big enough.

\begin{algorithm}[ht]
\caption{Seeded Iterative Clustering}\label{alg:cap}
\begin{algorithmic}
\State $X \gets embeddings$ \Comment{Embeddings of patches}
\State $y \gets seeds$ \Comment{Sparse annotations}
\State $score_{prev} \gets 0$ \Comment{Initialize score}

\State $labels \gets SeededKMeans(n_{clusters}=2).fit(X, y)$ \Comment{Initialize labels}

\While{$score \leq score_{prev}$} \Comment{Run until defined score goes down}
\State $X \gets embeddings[labels==1]$ \Comment{Select data only in the postive class}
\State $y \gets seeds[labels==1]$
\If{$\{0,1\} \subset y$} \Comment{If there are still annotations of both classes}
\State $labels \gets SeededKMeans(n_{clusters}=2).fit(X, y)$ \Comment{Re-cluster positive cluster}
\Else{}
    \If{$\{1\} \subset y$} \Comment{If there are only negative annotations}
    \State $labels \gets KMeans(n_{clusters}=2).fit(X)$ \Comment{Perform conventional clustering}
    \EndIf{}
\EndIf{}
\State $score \gets metric(y, label)$  \Comment{Compare only sparse annotations}
\EndWhile{}
\end{algorithmic}
\end{algorithm}

\subsection{Feature space visualization}

Inspired by the spatial-omics field, an intuitive way of visualizing the embeddings is by reducing the dimensionality of the space to three dimensions using Uniform Manifold Approximation and Projection for Dimension Reduction (UMAP) \cite{mcinnes2018umap}. This 3D space can be mapped to the RGB space and visualized directly on top of the images as colored dots at the center of each patch in the image, as previously presented by \citeauthor{chelebian2021morphological} \cite{chelebian2021morphological} and shown in Figure \ref{fig1} (bottom-right). The code for doing this can be found at \href{https://github.com/eduardchelebian/histology-umap}{github.com/eduardchelebian/histology-umap}.

In the UMAP space, it is possible to manually select regions that are clustered together and see if they correspond to biologically relevant regions on the image. This approach however, is highly dependent on the dimensionality reduction algorithm and should be used only for qualitative visualization purposes.

\section{Experimental results}

\subsection{Dataset}
For this proof-of-concept we used the DigestPath challenge cancer segmentation dataset \cite{da2022digestpath}. It is an open database including 250 images of tissue from 93 positive WSI (i.e. with pixel-level annotations) slides in JPEG format which were used for the experimental results. The additional 410 images from 231 WSI negative slides were only used together with the positive slides when self-supervised pretraining was performed. All WSI are hematoxylin-eosin (H\&E) stained and scanned at $20\times$ resolution.

\subsection{Experiments}

In order to explore how different embedding spaces would affect the clustering performance, we chose four different feature extraction frameworks: (1) ResNet-18 pretrained on ImageNet (ResNet18), (2) ResNet-50 pretrained on ImageNet (ResNet50), (3) ResNet-18 pretrained on the dataset using SimCLR self-supervision \cite{chen2020simple} (SimCLR) and (4) ResNet-18 pretrained on 57 histopathology datasets using SimCLR self-supervision \cite{ciga2022self} (HistoSSL). 

To mimic the complete workflow where the pathologist would select small benign and cancerous regions, we chose to experiment by randomly selecting 1, 3 and 5 patches that were located inside the benign and cancerous masks, respectively. Knowing that this would probably underestimate the results, as the seeds may not be representative of what a pathologist would choose, we report the F1-score for the patch classification task of the best run per slide in a 5-fold experiment, to simulate the selection an expert would make. The ground-truth coordinates are generated by selecting the patches that are at least 50\% in a cancer region. 

\subsection{Results}

\begin{figure}[ht]
  \centering
    \includegraphics[width=1\linewidth]{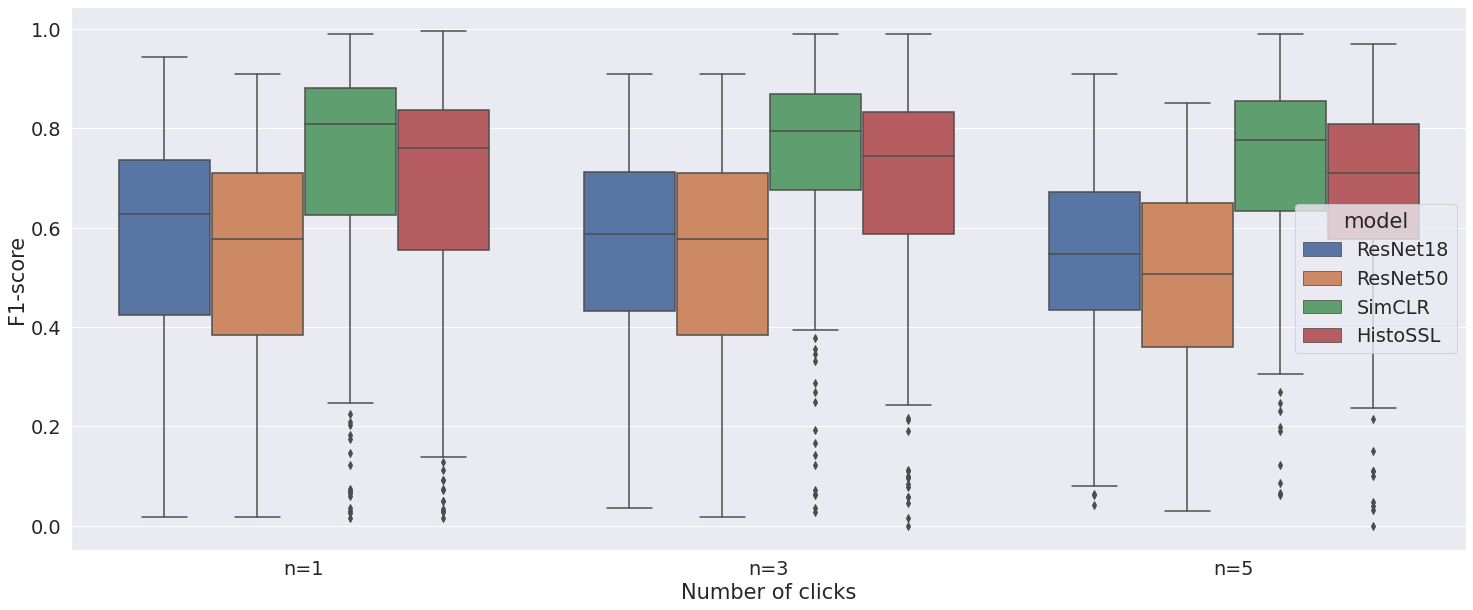}
  \caption{F1-score for the different models with varying number of simulated clicked patches.}
      \label{fig2}
\end{figure}

\begin{figure}[ht]
  \centering
    \includegraphics[width=1\linewidth]{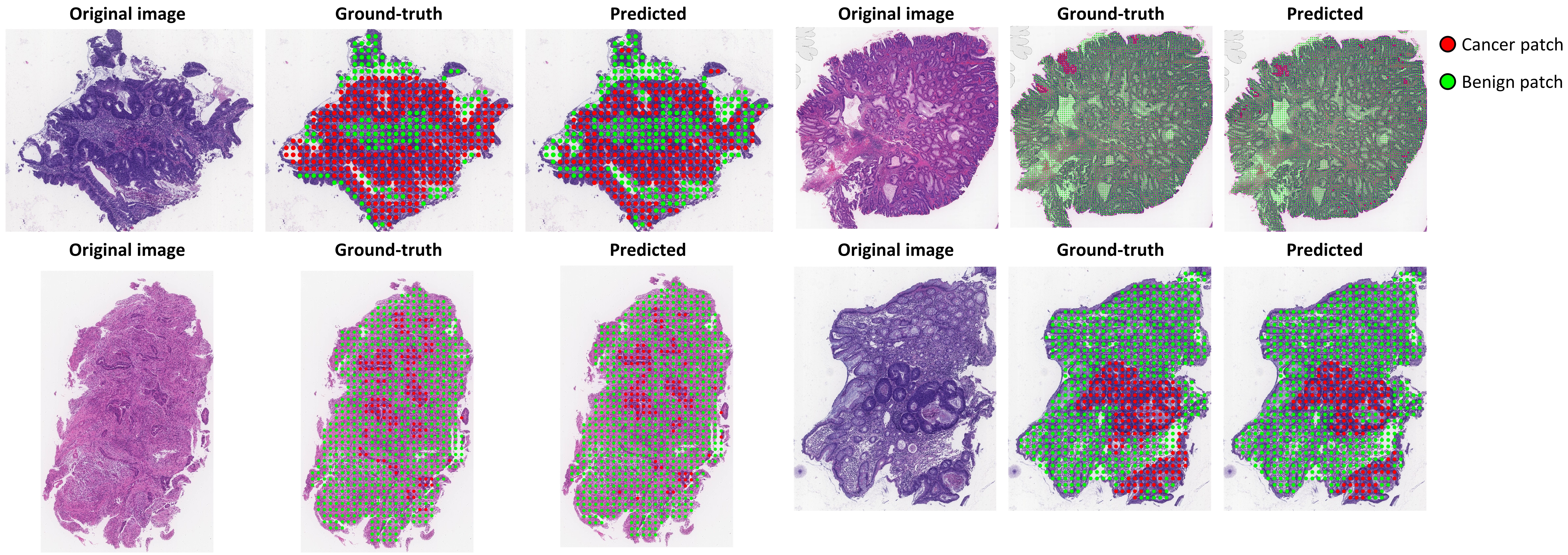}
  \caption{Examples of predictions using seeded iterative clustering. Visualizations created using TissUUmaps \cite{pielawski2022tissuumaps}.}
      \label{fig3}
\end{figure}

The results are summarized in Figure \ref{fig2}. The highest per-slide F1-score for the ImageNet pretrained models are very similar regardless of the architecure, $0.57 \pm 0.22$ for ResNet18 and $0.54 \pm 0.21$ for ResNet50, both lower than the self-supervised approaches. There seems to be some benefit for pretraining (SimCLR) on the dataset at hand ($0.74 \pm 0.23$), but this may not be always possible, so it is very promising that a publicly available model (HistoSSL) also achieves higher results ($0.68 \pm 0.22$). 
The method itself shows high classification variability, although the worse performing slides are usually the ones that have low positive patch ratio, that is, when the lesion is very small compared to the whole tissue, such as Figure \ref{fig3} (upper right). There is no clear difference from having more seeds, except when it comes to the standard deviation. As they are only used for initialization it seems that if the feature space is descriptive enough, adequately selected patches are sufficient for a good clustering result.

\section{Discussion}
We presented seeded iterative clustering (SIC) as a way of leveraging latent representations of neural networks to speed-up the time-consuming process of manual annotation in histopathology. From an annotation tool point of view, the local clustering module takes seconds, even on a normal computer. The embedding time depends on the amount of patches or coordinates one wants. However, contrary to previous work, this can be run remotely without the pathologist interaction and does not need to be recalculated at any point. Additionally, performing the clustering per-slide is more robust to domain shift artifacts.

From a technical point of view, it allowed to compare the representations from different networks. When performing transfer learning, there is a separate effect of the feature reuse and learning low-level statistics from the data \cite{neyshabur2020being}. Retraining, then, may not be ideal to explore the quality of the features. This is precisely why we find that ImageNet pretrained ResNet off-the-shelf are not comparable to using self-supervised pretraining with SimCLR.

Work in progress includes validating this approach on other datasets and developing it into a full interactive tool called in which pathologists select a handful of patches and the method proposes clusters for the rest. Additionally, the method needs to be compared against other approaches in order to confirm its feasibility. Finally, extending this approach in a multi-class problem (e.g. cancer, benign and stroma regions, instead of binary), would provide further insights into how the feature space generated by the different models as well as increase the usability of the tool. 

\section*{Broader Impact}

The uncoupling of the feature extraction and clustering modules could allow for the remote generation of embeddings that can then be shared with the community. This would allow for groups with less computational resources to work with the data and circumvent the potential privacy issues for sharing medical images, sharing instead the latent representation matrices per coordinate.

\begin{ack}
This research was funded by the European Research Council via ERC Consolidator grant CoG 682810 to C.W. and the SciLifeLab BioImage Informatics Facility.
\end{ack}

\medskip

\small

\printbibliography

\end{document}